\documentclass[conference]{IEEEtran}
\IEEEoverridecommandlockouts
\usepackage{cite}
\usepackage{amsmath,amssymb,amsfonts}
\usepackage{algorithmic}
\usepackage{graphicx}
\usepackage{textcomp}
\usepackage{xcolor}

\usepackage{xspace}
\usepackage{bm}
\usepackage{multirow}
\usepackage{booktabs}
\usepackage{tabularx}
\usepackage{dsfont}
\usepackage{physics}

\newcommand{\method}{{EITS}\xspace}

\def\BibTeX{{\rm B\kern-.05em{\sc i\kern-.025em b}\kern-.08em
    T\kern-.1667em\lower.7ex\hbox{E}\kern-.125emX}}
\begin{document}

%\title{
%Learning to Explore Informative Trajectories and Samples for %Embodied Perception
%}
% \vspace{6.3mm}  \LARGE \bf 
\title{\vspace{5mm}\LARGE \bf Learning to Explore Informative Trajectories and Samples for Embodied Perception}

\author{\IEEEauthorblockN{Ya Jing, Tao Kong}
\IEEEauthorblockA{ByteDance Research \\
\{jingya, kongtao\}@bytedance.com}
%\and
%\IEEEauthorblockN{2\textsuperscript{nd} Tao Kong}
%\IEEEauthorblockA{\textit{Bytedance} \\
%taokongcn@gmail.com}
}

\maketitle

\begin{abstract}
We are witnessing significant progress on perception models, specifically those trained on large-scale internet images. However, efficiently generalizing these perception models to unseen embodied tasks is insufficiently studied, which will help various relevant applications (e.g., home robots).
Unlike \textit{static} perception methods trained on pre-collected images, the embodied agent can move around in the environment and obtain images of objects from any viewpoints. 
Therefore, efficiently learning the exploration policy and collection method to gather informative training samples is the key to this task.
To do this, we first build a 3D semantic distribution map to train the exploration policy self-supervised by introducing the semantic distribution disagreement and the semantic distribution uncertainty rewards. 
Note that the map is generated from multi-view observations and can weaken the impact of misidentification from an unfamiliar viewpoint.
Our agent is then encouraged to explore the objects with different semantic distributions across viewpoints, or uncertain semantic distributions. 
With the explored informative trajectories, we propose to select hard samples on trajectories based on the semantic distribution uncertainty to reduce unnecessary observations that can be correctly identified. 
Experiments show that the perception model fine-tuned with our method outperforms the baselines trained with other exploration policies. Further, we demonstrate the robustness of our method in real-robot experiments. 
\end{abstract}

\begin{IEEEkeywords}
Embodied Perception, Trajectory Exploration, Hard Sample Selection
\end{IEEEkeywords}

\section{Introduction}
\label{sec:intro}
Pre-training on large-scale datasets to build reusable models has drawn great attention in recent years, e.g., the deep visual models \cite{he2016deep} pre-trained on ImageNet \cite{krizhevsky2012imagenet} can be reused for detection \cite{he2017mask,wang2020solo}, and pre-trained language model like BERT \cite{devlin2018bert} can be used for image-text retrieval \cite{chen2020uniter}. To better adapt to downstream tasks, many researchers focus on fine-tuning models on small-scale task-related datasets \cite{sun2019fine,wang2021list}. 
However, generalizing the perception model pre-trained on large-scale internet images to embodied tasks is insufficiently studied, which will help various relevant applications (e.g., home robots). In order to use as few annotations as possible, efficiently collecting training data in embodied scenes becomes the main challenge.

%Since the training images are collected from the Internet, these pre-trained models can successfully classify and segment objects in Internet images. 
%To generalize these pre-trained models to human-like robotic agent more efficiently instead of training from scratch, we need some domain knowledge about the new environment. Advances in active visual learning \cite{misra2018learning,sener2017active} have focused on efficient data collection techniques that can be used to well address this problem. 

Different from visual learning based on \textit{static} data (e.g., images), the embodied agent can \textit{move} around and interact with 3D environment. Therefore, efficiently collecting training samples means learning an exploration policy to encourage the agent to explore the areas where the pre-trained model performs poorly. Since the ground-truth labels in scenes are unavailable, the underlying spatial-temporal continuity in the 3D world can be used self-supervised. To use the consistency in semantic predictions, the previous method \cite{chaplot2020semantic} proposes a semantic curiosity policy, which explores inconsistent labeling of the same object by the perception model. When an exploration trajectory is learned, all observations on this trajectory are collected for labeling to fine-tune the pre-trained perception model. 
Despite the advance, this method utilizes a fuzzy inconsistency estimation (i.e., projecting multiple objects at different heights to the same location in a 2D map). 
%the uncertainty of the predicted semantic distribution that reflects the recognition performance of the pre-trained perception model in the new environment 
In addition, the uncertainty of the predicted semantic distribution that reflects what the pre-trained perception model does not know in the new environment and the hard sample selection on the trajectory are ignored in \cite{chaplot2020semantic}.

\begin{figure}[t]
\centering
\includegraphics[width=\linewidth]{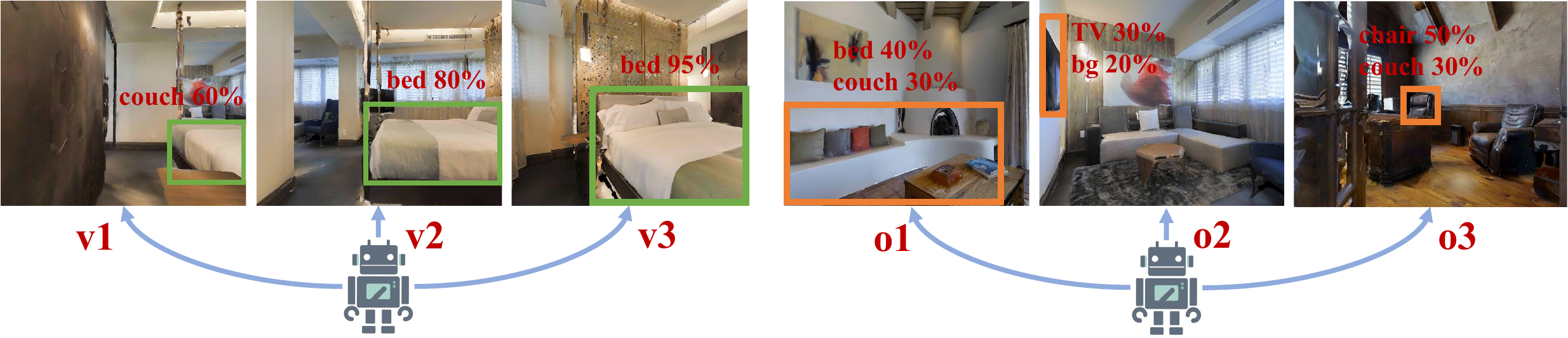}
\caption{The illustration of semantic distribution disagreement, e.g., the bed is recognized to different objects/distributions across three viewpoints ($v1, v2, v3$), and semantic distribution uncertainty, e.g., the probability of couch in observation $o1$  being predicted as bed and couch is relatively close. bg means background.}
\label{fig:illu}
\end{figure}

\begin{figure*}[t]
\centering
\includegraphics[width=0.92\linewidth]{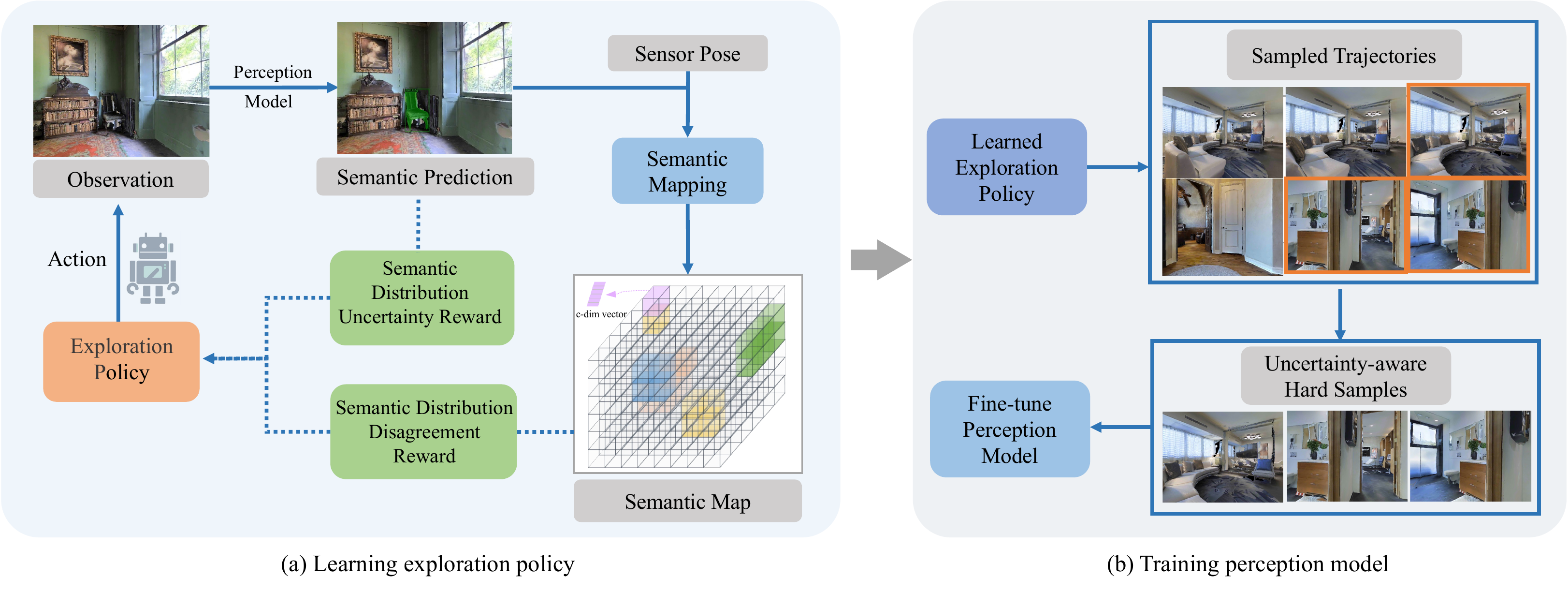}
\caption{The architecture of our proposed informative trajectory and sample exploration method. It contains two steps: the exploration policy aims to encourage the agent to explore the objects with semantic distribution disagreement or uncertainty, then the training stage aims at gathering hard samples on trajectories based on semantic distribution uncertainty to fine-tune the pre-trained model.}
\label{fig:main_model}
\end{figure*}

To solve these problems, we propose learning to Explore Informative Trajectories and Samples (\method) for embodied perception, as shown in Fig.~\ref{fig:main_model}. It consists of two steps: learning exploration policy and collecting hard samples to fine-tune the perception model.
During the exploration, our agent moves around and collects multi-view observations fused by Exponential Moving Average to generate a 3D semantic map, which weakens the impact of misidentification from an unfamiliar viewpoint.
The generated 3D map can be regarded as the pseudo ground truth due to the fusion of predicted results from different viewpoints. Unlike previous work \cite{chaplot2020semantic} that adopts predicted labels to build the semantic map, our work builds a predicted probabilistic distribution map, i.e., a 3D semantic distribution map. 
It can be used to constrain not only the predicted labels but also the predicted distributions, 
%to reduce the uncertainty of predictions, 
as shown in Fig.~\ref{fig:illu} (left).
Then one curiosity reward is measured by the semantic distribution disagreement between the semantic prediction of the current perspective and the generated 3D semantic distribution map. We also measure the uncertainty of the predicted semantic distribution as shown in Fig.~\ref{fig:illu} (right) as another curiosity reward. These two rewards are used together to learn the exploration policy by maximizing the disagreement and uncertainty of semantic predictions. Therefore, our agent can move to the areas where the 
semantic predictions are different from pseudo ground truth or the
probabilities being predicted as two categories are relatively close. 
%The probabilities of being predicted as two classes are relatively close.
%confidences belonging to two categories are relatively close.. 
After obtaining the exploration strategy, we gather indistinguishable hard samples on each trajectory for the subsequent training. The selection method is the same as the uncertainty of semantic distribution. 
The gathered data is labeled and utilized to fine-tune the pre-trained perception model, making it generalized well to new environments. We also show that the perception model could be sustainably improved by our explore-finetune process back and forth.
The method is evaluated on a challenging Matterport dataset \cite{chang2017matterport3d} and our real-robot environment. Experimental results show that our \method approach outperforms the state-of-the-art methods.

The contributions of this paper can be summarized as threefold. (1) We propose a novel informative trajectory exploration method for embodied perception by measuring the semantic distribution disagreement across viewpoints and the uncertainty of each observation. In addition, we are probably the first to exploit the uncertainty over semantic predictions to handle this task. 
% measured by 3D semantic distribution map
(2) The hard samples on the trajectory selected by the semantic distribution uncertainty further sieves the observations recognized well by the pre-trained model, which can enhance the performance better when fine-tuning the pre-trained model. (3) The proposed method achieves the best result on the challenging Matterport dataset. 
%Extensive ablation studies verify the effectiveness of each component in the \method. 

\section{Related Work}
\label{sec:related}
%In this section, we briefly introduce the related work about prior studies on active perception and learning, semantic mapping, as well as some embodied tasks. 

\subsection{Robot Perception Learning}
Visual perception is a crucial function of a robot. Some works \cite{chaplot2020object,zhu2022navigating} directly utilize the perception model pre-trained on COCO \cite{lin2014microsoft} images to perform object goal navigation. To improve the performance in embodied tasks, some researchers \cite{jayaraman2016look,yang2019embodied,yeung2016end} focus on learning a policy to directly improve metrics of interest end-to-end at test time by obtaining information about the environment. Unlike them, we aim to explore informative samples self-supervised to better fine-tune the pre-trained perception model. 
%Active learning \cite{yang2018visual,pathak2018learning,jayaraman2018end} also aims to actively select a subset of unlabeled images for labeling to improve its model, but it is based on pre-collected datasets. 
%It has wide applications, e.g., object detection \cite{jayaraman2018end,vijayanarasimhan2014large,yang2018visual}, instance segmentation \cite{pathak2018learning} and medical image analysis \cite{kuo2018cost}
The exploration in reinforcement learning \cite{eysenbach2018diversity,pathak2017curiosity,pathak2019self} also aims to maximize an intrinsic reward function to encourage the agent to seek previously unseen or poorly understood parts of the environment. 
%In addition, Pathak et al. \cite{pathak2019self} measure the consistency among multiple models. 
Different from them, we compute the reward function by multi-view consistency in semantics. 
The active interactive imitation learning \cite{hoque2021thriftydagger, menda2019ensembledagger, hoque2021lazydagger} are also related to our work in disagreement and uncertainty measuring to decide whether to request a label from the human. However, our agent does not require human intervention when learning the exploration policy, and the exploration purpose is to improve the perception model. 
Recently, Chaplot et al. \cite{chaplot2020semantic} measure the semantic curiosity to learn the exploration policy for embodied perception. But they ignore the uncertainty over semantic predictions and hard sample selection on the learned trajectory. Besides, some works \cite{chaplot2021seal, fang2020move} attempt to learn both exploration and perception utilizing pseudo labels in a completely self-supervised manner without requiring any extra labels. In this paper, we propose effectively generalizing the pre-trained perception model to embodied tasks, where informative trajectories and samples are gathered by utilizing a 3D semantic distribution map to measure the semantic distribution disagreement and the semantic distribution uncertainty. Then the gathered data is labeled to fine-tune the perception model.

\subsection{Semantic Mapping}
3D mapping aiming to reconstruct a dense map of the environment has achieved great advances in recent years. Fuentes-Pacheco et al. \cite{fuentes2015visual} do a very detailed survey. Researchers also consider adding semantic information to the 3D map \cite{chaplot2021seal}. Similar to them, we adopt the same setting and learn 3D semantic mapping by differentiable projection operations. In this paper, we propose a 3D semantic distribution map, which is used to learn the exploration policy.
% constrain the semantic distributions to learn exploration policy. 

\subsection{Embodied Task}
Embodied agents can move around and interact with the surrounding environment. Many environments are photo-realistic reconstructions of indoor \cite{chang2017matterport3d,xia2018gibson} and outdoor \cite{dosovitskiy2017carla,geiger2013vision} scenes, where the ground-truth labels for objects are also provided. Recently, many researchers have used these simulated environments in visual navigation \cite{chaplot2020object,gupta2017cognitive}, visual question answering \cite{das2018embodied} and visual exploration \cite{chaplot2020learning}. Visual navigation usually involves point/object goal navigation \cite{chaplot2020object} and vision-and-language navigation \cite{anderson2018vision} where the path to the goal is described in natural language. 
Visual question answering \cite{das2018embodied} should intelligently navigate to explore the environment, gather necessary visual information, and then answer the question. 
% Recent works explicitly train an end-to-end RL policy to maximize the explored area in visual exploration \cite{chaplot2020learning}. 
Unlike them, our agent aims to gather data for labeling to generalize the pre-trained perception model to unseen environments efficiently. 

\section{Approach}
\label{sec:method}

We aim to train an embodied agent with a perception model pre-trained on internet images to explore informative trajectories and samples effectively. Then the perception model fine-tuned on the gathered data can generalize well to a new environment. As shown in Fig.~\ref{fig:main_model}, our proposed method consists of two main parts. The exploration part aims to learn the active movement of an agent to obtain informative trajectories via semantic distribution disagreement and semantic distribution uncertainty self-supervised. %utilizing a 3D semantic distribution map. 
Then we take advantage of the semantic distribution uncertainty to collect hard samples on the learned trajectory. After images are collected and semantically labeled, we fine-tune the perception model on these images.
%at the perception learning stage. 

\subsection{3D Semantic Distribution Mapping}

Note that for each time step $t$, our agent's observation space consists of an RGB observation $I_{t} \in \mathbb R^{3 \times W_{I} \times H_{I}}$, a depth observation $D_{t} \in \mathbb R^{W_{I} \times H_{I}}$, and a 3-DOF pose sensor $x_{t} \in \mathbb R^{3}$ which denotes the $x$-$y$ coordinates and the orientation of the agent. The agent has three discrete actions: \texttt{move forward}, \texttt{turn left} and \texttt{turn right}. 

The easiest way to associate semantic predictions across frames on a trajectory is to project the predictions on the top-down view to build a 2D semantic map as \cite{chaplot2020semantic}. 
However, due to the embodied agent moving in a 3D environment, the height information is lost when projecting the predictions onto a 2D map. These will result in projecting multiple objects at different heights to the same location, e.g., if a potted plant is on the table, the potted plant and table will be projected to the same location. Therefore, the noise will be generated when calculating the disagreement across different viewpoints. 
In this paper, we utilize the 3D semantic distribution map to measure the semantic distribution disagreement.
%, which is a voxel-based representation. 
The semantic map $M$ is a 4D tensor of size $K \times L_{M} \times W_{M} \times H_{M}$, where $L_{M}$, $W_{M}$, $H_{M}$, denote the 3 spatial dimensions, and $K = C + 2$, where $C$ is the total number of semantic object categories. The first two channels in $K$ represent whether the corresponding voxel (x-y-z location) contains obstacles and is the explored area, respectively. The other channels denote the predicted semantic probability distribution among $C$ categories from the pre-trained perception model.
%The $P$ denotes the semantic probability distribution among $C$ categories from pre-trained perception model. 
The map is initialized with all zeros at the beginning of an episode, $M_0 = [0]^{K \times L_{M} \times W_{M} \times H_{M}}$. The agent always starts at the center of the map facing east at the beginning of the episode, $x_0 = (L_{M}/2, W_{M}/2, 0.0)$ same as \cite{chaplot2020object}. 

Fig.~\ref{fig:main_model} shows the 3D semantic mapping procedure at a time step. The agent takes action and then sees a new observation $I_{t}$. The pre-trained perception model (e.g., Mask RCNN \cite{he2017mask}) is adopted to predict the semantic categories of the objects seen in $I_{t}$, where the semantic prediction is a probability distribution among $C$ categories for each pixel. The depth observation $D_{t}$ is used to compute the point cloud. Each point in the point cloud is associated with the corresponding semantic prediction, which is then converted into 3D space using differentiable geometric transformations based on the agent pose to get the voxel representation. This voxel representation in the same location is aggregated over time using Exponential Moving Average to get the 3D semantic distribution map:
%\begin{equation}
   %M_{t} = M_{t-1}, t=1
%\end{equation}
%\begin{equation}
   %M_{t} = \lambda * M_{t-1} + (1-\lambda) * m_{t}, t>1
%\end{equation}
\begin{equation}
\begin{split}
   M_{t} = \left \{
\begin{array}{ll}
   M_{t-1}, & t=1 \\
   \lambda * M_{t-1} + (1-\lambda) * m_{t}, & t>1 \\
\end{array}
\right.
\end{split}
\end{equation}
where $m_{t}$ means the voxel representation at time step $t$ and $\lambda$ aims to control the relative importance of $M_{t-1}$ and $m_{t}$. The map can integrate the predicted semantics of the same object from different viewpoints to alleviate the misrecognition caused by the unfamiliar viewpoint. Therefore, the map representation can be used as pseudo ground truth labels of objects in the scene.

\subsection{Exploring Informative Trajectory}

The goal of exploration policy $a_{t} = \pi (I_{t}, \theta)$ is exploring objects that are poorly identified by the current perception model based on the observation $I_{t}$, where $a_{t}$ means the action and $\theta$ represents the parameters of the policy model. 
Hence, we can collect valuable observations in the explored areas to fine-tune the perception model. We propose two novel distribution-based rewards to train the exploration policy by maximizing the disagreement and uncertainty during moving.

The semantic distribution disagreement reward is defined as the Kullback-Leibler divergence between the current prediction and the 3D semantic distribution map, which encourages the agent to explore the objects with different semantic distributions across viewpoints:
%not only the new objects but also the objects with different predictions across viewpoints:
\begin{equation}
   r_d = KL(m_{t}, M_{t-1}).
\end{equation}
Unlike semantic curiosity \cite{chaplot2020semantic} which maximizes the label inconsistency based on the 2D semantic map, our semantic distribution disagreement aims to explore the objects with different distributions from the 3D semantic distribution map. 

In addition, we propose a semantic distribution uncertainty reward $r_u$ to explore the objects whose predicted probabilities belonging to two categories are relatively close, as Eq.~\ref{4} explains. 
\begin{equation}
\begin{split}
   r_u = \left \{
\begin{array}{ll}
   1, & u>\delta \\
   0, & u<\delta \\
\end{array}
\right.
\end{split}
\end{equation}
To train the policy, we first input the semantic map to a global exploration policy to select a long-term goal (i.e., an x-y coordinate of the map). Then a deterministic Fast Marching Method \cite{sethian1996fast} is used for path planning, which uses low-level navigation actions to achieve the goal. We sample the long-term goal every 25 local steps, same as \cite{chaplot2020object} to reduce the time horizon for exploration in reinforcement learning. The Proximal Policy Optimization (PPO) %\cite{schulman2017proximal} 
is used to train the policy. 

\subsection{Efficient Sample Selection and Continue Training}

After obtaining the trajectory, the easiest way is to label all observations on the trajectory. Although the trained exploration policy can find more objects with inconsistent and uncertain predictions, there are still many observations that the pre-trained model can accurately identify. To efficiently fine-tune the perception model, we propose a sample selection method by measuring the uncertainty $u$ of the semantic distribution:
%The uncertainty estimation $u$ is computed as:
\begin{equation}
\label{4}
   u = Second_{max}(P_{i}),
\end{equation}
where $P_{i} \in {\mathbb R} ^{C}$ is the predicted class probability of $i$th object in a single image, the $Second_{max}$ means the second largest score in $\{ p_{i}^{0}, p_{i}^{1}, ..., p_{i}^{C-1} \}$. If $u$ is larger than threshold $\delta$, we select the corresponding image. Considering that the semantic distribution disagreement relies heavily on multi-view observations in the trajectory will reduce the efficiency of selection, and thus it is not utilized to select hard samples. 

We label the selected images and use them to fine-tune the perception model. 

\newcolumntype{Z}{p{1.2cm}<{\centering}}
\newcolumntype{W}{p{1.5cm}<{\centering}}
\begin{table*}[t]
\centering
\caption{Comparison with the state-of-the-art methods for object detection (Bbox) and instance segmentation (Segm) using AP50 as the metric. n means the exploration policy is progressively trained for n times. \label{tab:sota}}
\begin{tabularx}{\linewidth}{Z|X|ZZWZZZ|Z}
\toprule
{Task}&{Method}&{Chair}&{Couch}&{Potted Plant}&{Bed}&{Toilet}&{Tv}&{Average} \\
\hline
{}&{Pre-trained}&{21.05}&{25.23}&{22.58}&{24.24}&{22.62}&{29.67}&{24.23}\\
{}&{Re-trained}&{23.77}&{27.36}&{26.20}&{25.21}&{24.82}&{34.48}&{26.97}\\
{}&{Random}&{29.98}&{31.65}&{23.91}&{28.66}&{31.78}&{40.44}&{31.07}\\
{Bbox}&{Active Neural SLAM \cite{chaplot2020learning}}&{32.02}&{32.74}&{31.94}&{30.31}&{26.30}&{38.68}&{32.00}\\
{}&{Semantic Curiosity \cite{chaplot2020semantic}}&{33.51}&{33.11}&{32.91}&{29.57}&{25.76}&{39.97}&{32.46}\\
{}&{Ours (n=1)}&{33.57}&{34.36}&{32.79}&{31.54}&{28.38}&{43.81}&{34.07}\\
{}&{Ours (n=3)}&{33.34}&{34.48}&{35.28}&{32.12}&{31.87}&{43.11}&{\bf{35.03}}\\
\hline
\hline
{}&{Pre-trained}&{12.72}&{22.98}&{16.71}&{23.82}&{23.85}&{29.75}&{21.64}\\
{}&{Re-trained}&{14.99}&{24.68}&{18.36}&{24.32}&{25.15}&{34.23}&{23.62}\\
{}&{Random}&{18.22}&{27.25}&{8.82}&{28.19}&{29.08}&{39.39}&{25.16}\\
{Segm}&{Active Neural SLAM \cite{chaplot2020learning}}&{17.89}&{29.24}&{15.22}&{29.66}&{27.29}&{38.61}&{26.32}\\
{}&{Semantic Curiosity \cite{chaplot2020semantic}}&{18.18}&{30.06}&{18.39 }&{29.03}&{26.70}&{40.01}&{27.06}\\
{}&{Ours (n=1)}&{19.18}&{30.14}&{15.56}&{31.03}&{28.19}&{43.43}&{27.92}\\
{}&{Ours (n=3)}&{19.28}&{30.13}&{16.22}&{31.27}&{28.92}&{44.76}&{\bf{28.42}}\\
\bottomrule
\end{tabularx}
\end{table*}

\section{Experiments}
\label{sec:exp}
\subsection{Implementation details}

We use the Matterport3D \cite{chang2017matterport3d} dataset with Habitat simulator \cite{savva2019habitat} in our main experiments. The scenes in the Matterport3D dataset are 3D reconstructions of real-world environments, split into a training set (54 scenes) and a test set (10 scenes). We assume that the perfect agent pose and depth image can be obtained in our setup. 

The exploration policy consists of convolutional layers followed by fully connected layers. 
%In addition to the semantic map, we also input the agent orientation to the policy.
The pre-trained Mask RCNN is frozen while training the exploration policy. We use the PPO with a time horizon of 20 steps, 8 mini-batches, and 4 epochs in each PPO update to train the policy. The reward, entropy, and value loss coefficients are set to 0.02, 0.001, and 0.5, respectively. We use Adam optimizer with a learning rate of $2.5 \times 10^{-5}$. The maximum number of steps in each episode is 500. The $\lambda$ and $\delta$ are experimentally set to 0.3 and 0.1, respectively. To fairly compare with previous methods, we set the number of training steps to 500k in all experiments.

We pre-train a Mask-RCNN model with FPN \cite{lin2017feature} using ResNet-50 as the backbone on the COCO \cite{lin2014microsoft} dataset labeled with 6 overlapping categories with the Matterport3D, i.e., ‘chair’, ‘couch’, ‘potted plant’, ‘bed’, ‘toilet’ and ‘tv’. Then we fine-tune this model on the gathered samples with a fixed learning rate of 0.001. All other hyper-parameters are set to default settings in Detectron2 \cite{wu2019detectron2}. We randomly collect the samples in test scenes of different episodes to evaluate the final perception model. The AP50 score is adopted as the evaluation metric, which is the average precision with at least $50\%$ IOU. 

We further deploy our method to a real robot. Our robot is equipped with a Kinect V2 camera, a 2D LiDAR, and an onboard computer (with an Intel i5- 7500T CPU and an NVIDIA GeForce GTX 1060 GPU). Note that the LiDAR is only used with wheel odometers to perform localization. We test our method in a built 60$m^{2}$ house with a dining room, a living room, and a bedroom. 
%In real robots, we use the sensor fusion (i.e., 2D LiDAR and wheel odometer) to reduce the self-positioning errors.

%We also show that our method could also guide the real robot to find informative trajectories and samples. 

% 两类无法区分：RPN：0.2, 0.3, 0.4, 0.5 thres: 0.1, 0.2, 0.3 
\newcolumntype{Z}{p{1.1cm}<{\centering}}
\newcolumntype{W}{p{1.5cm}<{\centering}}
\begin{table*}[t]
\centering
\caption{Effects of setting different thresholds in hard sample selection on object detection task. \label{tab:sam}}
\begin{tabularx}{\linewidth}{X<{\centering}|X<{\centering}|ZZWZZZ|W}
\toprule
{Method}&{Training Image}&{Chair}&{Couch}&{Potted Plant}&{Bed}&{Toilet}&{Tv}&{Average} \\
\hline
{$\delta$ = 0.1}&{20k}&{33.57}&{34.36}&{32.79}&{31.54}&{28.38}&{43.81}&{34.07}\\
{$\delta$ = 0.2}&{13k}&{33.38}&{34.61}&{31.34}&{31.84}&{26.24}&{41.64}&{33.18}\\
{$\delta$ = 0.3}&{9k}&{32.79}&{35.03}&{31.10}&{31.81}&{22.83}&{41.11}&{32.44}\\
\bottomrule
\end{tabularx}
\end{table*}

\subsection{Main Results}
\subsubsection{Simulation Environment}
To demonstrate the effectiveness of our method, we compare our fine-tuned object detection and instance segmentation results with the state-of-the-art methods as shown in Tab.~\ref{tab:sota}. Note that these methods all use around 20k training images. Pre-trained means the perception model was pre-trained on the raw COCO dataset. Re-trained means we re-train the pre-trained model utilizing COCO dataset labeled with 6 overlapping categories with the Matterport3D. Random is a baseline exploration policy that samples actions randomly. It can be seen that our model achieves the best performance and can further improve the performance when progressively training the exploration strategy three times based on the latest fine-tuned perception model. 

%%%%%%%%%%%%%%%%%%%%%%%%%%%%%%%%%
\iffalse
\begin{table}[h]
\caption{\color{red}Ablation studies on the object detection task. SDD and CDU means the semantic distribution disagreement reward and the class distribution uncertainty reward in trajectory exploration, respectively. HSM means the hard sample mining. SC means the Semantic Curiosity \cite{chaplot2020semantic}.}
\label{tab:abl}
%\resizebox{\linewidth}{!}{
\setlength{\tabcolsep}{2.8mm}{
%\centering
\begin{tabular}{l|cccccc|c}
\hline
{Method}&{Chair}&{Couch}&{Potted Plant}&{Bed}&{Toilet}&{Tv}&{Average} \\
\hline
{\color{red}Ours w/o SDD}&{32.08}&{34.51}&{32.05}&{29.91}&{27.95}&{42.76}&{33.21} \\
{Ours w/o CDU}&{33.49}&{34.39}&{32.95}&{31.19}&{27.31}&{42.18}&{33.59}\\
{Ours w/o HSM}&{32.44}&{33.69}&{32.22}&{30.85}&{27.67}&{41.78}&{33.11}\\
{SC + HSM}&{33.87}&{33.52}&{32.69}&{31.08}&{27.93}&{41.08}&{33.36}\\
{Ours}&{33.57}&{34.36}&{32.79}&{31.54}&{28.38}&{43.81}&{\bf{34.07}}\\
\hline
\end{tabular}
}
\end{table}
\fi
%%%%%%%%%%%%%%%%%%%%%%%%%%%%%%%%%

\newcolumntype{Z}{p{0.4cm}<{\centering}}
\newcolumntype{W}{p{0.7cm}<{\centering}}
\newcolumntype{M}{p{0.5cm}<{\centering}}
\begin{table}[t]
\centering
\caption{Ablation studies on the object detection task. SDD and SDU means the semantic distribution disagreement reward and the semantic distribution uncertainty reward in trajectory exploration, respectively. HSS means the hard sample selection. SC means the Semantic Curiosity \cite{chaplot2020semantic}. \label{tab:abl}}
\begin{tabularx}{\linewidth}{X<{\centering}|ZZZZZM|W}
\toprule
{Method}&{Chair}&{Couch}&{Potted Plant}
&{Bed}&{Toilet}&{Tv}&{Average} \\
\hline
{Ours w/o SDD}&{32.08}&{34.51}&{32.05}&{29.91}&{27.95}&{42.76}&{33.21} \\
{Ours w/o SDU}&{33.49}&{34.39}&{32.95}&{31.19}&{27.31}&{42.18}&{33.59}\\
{Ours w/o HSS}&{32.44}&{33.69}&{32.22}&{30.85}&{27.67}&{41.78}&{33.11}\\
{SC + HSS}&{33.87}&{33.52}&{32.69}&{31.08}&{27.93}&{41.08}&{33.36}\\
{Ours}&{33.57}&{34.36}&{32.79}&{31.54}&{28.38}&{43.81}&{\bf{34.07}}\\
\bottomrule
\end{tabularx}
\end{table}

Specifically, compared with the pre-trained model, our fine-tuned model gives 10.80$\%$ AP50 gains on the box detection metric.
Compared with the previous best competitor Semantic Curiosity \cite{chaplot2020semantic} which rewards trajectories with inconsistent labeling behavior and encourages the embodied agent to explore such areas, our model significantly outperforms it by 2.57$\%$ absolute AP50 point on object box detection and 1.36$\%$ on instance segmentation. The improved performances over the best competitor indicate that our proposed informative trajectory exploration and hard sample selection method is very effective for this task. 
%Compared with the Active Neural SLAM \cite{chaplot2020learning} which aims to maximize the total explored areas, our model achieves much better performances by 2.07$\%$, demonstrating the effectiveness of our data gathering policy. 
In addition, we can see that our method is more friendly to instances with simple shapes, e.g., Bed and Tv. These instance's shapes are easier to be reconstructed through 3D mapping. Objects with much more complicated shapes, e.g., Potted Plant, are more likely to involve mapping errors, which in turn decreases the performance of instance segmentation.

\subsubsection{Real Robot}
We also deploy our learned exploration 
policy on a real robot to explore informative trajectories and hard samples in an unseen environment. In practice, we gather 170 hard samples for fine-tuning the pre-trained model and an additional 50 randomly collected samples for validation, with an average of 4 objects in each image. Benefiting from the gathered informative images, the fine-tuned perception 
model can improve the detection and segmentation performances from 79.1\% AP50 and 76.7\% AP50 to 97.3\% AP50 and 96.1\% AP50, respectively.

\subsection{Ablation Analysis}

Our method comprises two modules: informative trajectory exploration and hard sample selection. To investigate these two components, we perform a set of ablation studies with $n$ = 1 for simplicity, as shown in Tab.~\ref{tab:abl}.

We first investigate the importance of rewarding semantic distribution disagreement across viewpoints and semantic distribution uncertainty to explore the trajectory. 
It can be seen that the AP50 accuracy on object detection drops 0.71$\%$ (SC+HSS vs. Ours) by replacing our exploration policy as SC.
The exploration module proves the effectiveness of learning informative trajectories for subsequent sample selection. Then we investigate the importance of semantic distribution uncertainty based hard sample selection by removing it (Ours w/o HSS).
The AP50 accuracy on object detection drops 0.96$\%$, demonstrating that selecting hard samples enhances the perception results. In addition, by comparing the results between Ours and Ours w/o SDD, Ours and Ours w/o SDU (semantic distribution disagreement and uncertainty in informative trajectory exploration), we can find that utilizing SDD and SDU can generate more effective trajectories. 

%Experimental results show that 'Ours w/o CPU' model selects more Couch and Potted Plant categories to be labeled (16470 VS 15519; 11246 VS 10743), indicating that these two categories of objects have prediction uncertainty at the same time when having prediction disagreement. 

\newcolumntype{Z}{p{0.5cm}<{\centering}}
\newcolumntype{W}{p{0.8cm}<{\centering}}
\begin{table}[t]
\centering
\caption{Effects of progressively training the exploration policy for $n$ times on the object detection task. \label{tab:multi}}
\begin{tabularx}{\linewidth}{X<{\centering}|ZZZZZZ|W}
\toprule
{Method}&{Chair}&{Couch}&{Potted Plant}&{Bed}&{Toilet}&{Tv}&{Average} \\
\hline
{n = 1}&{33.57}&{34.36}&{32.79}&{31.54}&{28.38}&{43.81}&{34.07}\\
{n = 2}&{32.18}&{33.32}&{36.06}&{31.38}&{30.81}&{44.53}&{34.71}\\
{n = 3}&{33.34}&{34.48}&{35.28}&{32.12}&{31.87}&{43.11}&{\bf{35.03}}\\
\bottomrule
\end{tabularx}
\end{table}

\begin{figure*}[t]
\centering
\includegraphics[width=\linewidth]{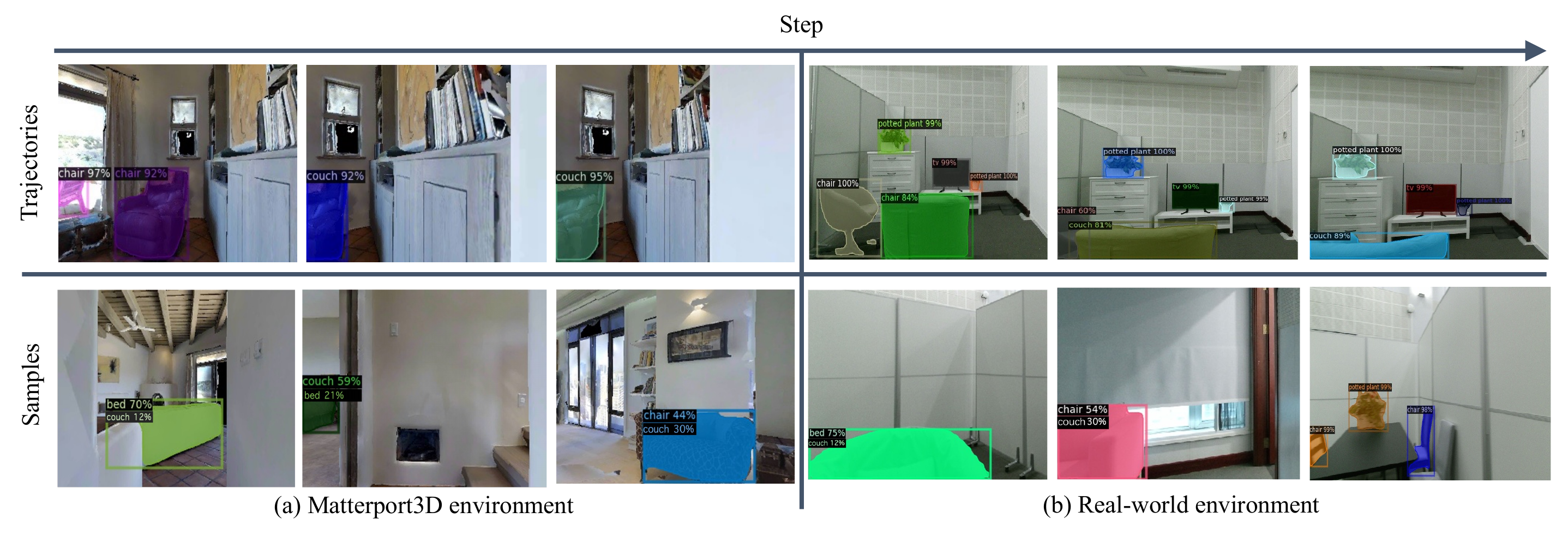}
\caption{Qualitative examples of learned trajectories and sampled images from the Matterport3D environment and the real robot. The first row shows the explored informative trajectories trained by semantic distribution disagreement and uncertainty rewards. The second row shows the gathered hard images by semantic distribution uncertainty estimation.}
\label{fig:trajectory}
\end{figure*}

\begin{figure}[t]
\centering
\includegraphics[width=\linewidth]{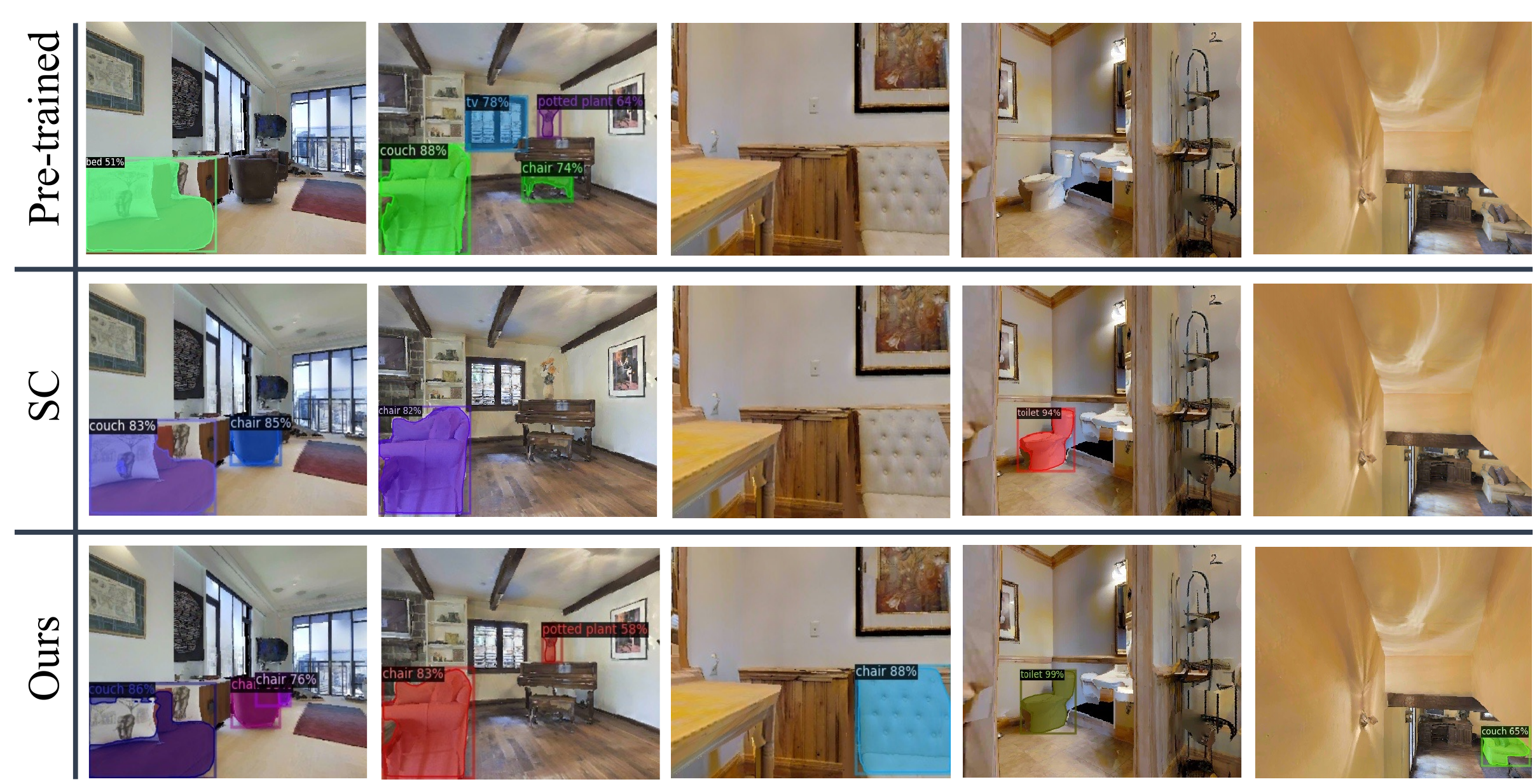}
\caption{Qualitative examples of instance segmentation by different models.}
\label{fig:seg}
\end{figure}

We compare the effectiveness of setting different thresholds $\delta$ in hard sample selection as shown in Tab.~\ref{tab:sam}. In this experiment, we sample the images from explored trajectories with 6 episodes and fixed steps in each training scene, resulting in different numbers of sampled training images at different thresholds. We can find that decent performances can be achieved by training very few hard samples, which demonstrates the effectiveness of selecting hard samples. Tab.~\ref{tab:multi} shows the experimental results when progressively training the exploration policy multiple times based on the latest fine-tuned perception model. Note that they all use 20k training images.
%Considering the simplicity, we adopt $n=1$ in this paper. 

To exploit measures of uncertainty in semantic distributions, we utilize the entropy of categorical distribution (ECS) in place of the heuristic in Eq.~\ref{4} as shown in Tab.~\ref{tab:uncetain}. We experimentally set the threshold of entropy to 0.4. The improved performance indicates that the uncertainty between all distributions is more effective than between the two categories.

\newcolumntype{Z}{p{0.5cm}<{\centering}}
\newcolumntype{W}{p{0.8cm}<{\centering}}
\begin{table}[t]
\caption{Effects of utilizing the entropy of categorical distribution (ECS) in place of the heuristic in Eq.~\ref{4} to measure semantic distribution uncertainty on the object detection task. \label{tab:uncetain}}
\begin{tabularx}{\linewidth}{X<{\centering}|ZZZZZZ|W}
\toprule
{Method}&{Chair}&{Couch}&{Potted Plant}&{Bed}&{Toilet}&{Tv}&{Average} \\
\hline
{Ours}&{33.57}&{34.36}&{32.79}&{31.54}&{28.38}&{43.81}&{34.07}\\
{ECS}&{31.40}&{32.70}&{34.23}&{30.90}&{30.75}&{46.06}&{\bf{34.34}}\\
\bottomrule
\end{tabularx}
\end{table}

\subsection{Qualitative Results}

To verify whether the proposed exploration policy and hard sample selection method can obtain the observations with inconsistent or uncertain semantic distributions, we visualize the explored trajectories and sampled images from the Matterport3D dataset and real-world environment, as shown in Fig.~\ref{fig:trajectory}. We can see that our model is able to gather inconsistent and uncertain detections via semantic distribution disagreement and uncertainty estimation. 
For example, the couch is detected as different objects (chair/couch) or distributions from different viewpoints on the first row. Besides, the couch is detected as couch and chair with almost close scores on the second row. By collecting these observations that are poorly identified by the pre-trained perception model for labeling, the model can be fine-tuned better. 

Fig.~\ref{fig:seg} shows the segmentation masks obtained by three different models, i.e., pre-trained, Semantic Curiosity \cite{chaplot2020semantic}, and our \method, demonstrating our proposed method's benefits. As the figure shows, our generated segmentation masks have more obvious object shapes and finer outlines in the first column. Besides, our model, fine-tuned exclusively on hard samples, can detect the missed objects by the pre-trained and Semantic Curiosity \cite{chaplot2020semantic} models, as shown in the third and fifth columns. 

\section{Discussion and Limitations}
\label{sec:conclusion}
We propose to generalize the perception model pre-trained on internet images to the unseen 3D environments with as few annotations as possible. Therefore, efficiently learning the exploration policy and selection method to gather training samples is the key to this task. In this work, we propose a novel informative trajectory exploration method via semantic distribution disagreement and semantic distribution uncertainty. Then the uncertainty-based hard sample selection method is proposed to further reduce unnecessary observations that can be correctly identified. Extensive ablation studies verify the effectiveness of each component of our method.

Although our method is more efficient than previous works, there are still some limitations. Through exploring the informative trajectories and samples, we can efficiently generalize the pre-trained model to the embodied task, where labeling the
segmentation mask is still costly. The weakly-supervised methods (e.g., utilizing box annotations to train segmentation models) can be utilized to fine-tune the perception model in the future. In addition, we collect all samples before fine-tuning the perception model, which results in our perception model not being updated. In the future, we can explore updating the perception module when learning the exploration policy. 

%The training of exploration policy is based on the perception model, which will result in policy retraining when using a new perception model. 

\section{Acknowledgments}
\label{sec:acknowledgments}
We would like to thank Minzhao Zhu, Yifeng Li, Yuxi Liu, Tao Wang and Yunfei Liu for their help on the robot system, Hang Li for helpful feedback, and other colleagues at ByteDance AI Lab for support throughout this project.

\bibliographystyle{IEEEtran}
\bibliography{eihe}

%\clearpage
%\appendix 
%\label{sec:appendix}
%\input{070appendix.tex}

\end{document}